\documentclass[conference]{IEEEtran}
\usepackage{graphicx}
\usepackage{times}
\usepackage[numbers]{natbib}
\usepackage{multicol}
\usepackage{hyperref}

\usepackage[utf8]{inputenc} 
\usepackage{amsmath,amsfonts}
\usepackage{algorithm2e}
\usepackage{todonotes}

\usepackage{bmpsize}
\usepackage{subcaption}
\usepackage{authblk}

\usepackage{fancyhdr}
\begin{document}

\title{Harnessing Reinforcement Learning for Neural Motion Planning}
\author[1]{Tom Jurgenson}
\author[1]{Aviv Tamar}
\affil[1]{Faculty of Electrical Engineering, Technion}

\maketitle
\begin{abstract}
Motion planning is an essential component in most of today's robotic applications. 
In this work, we consider the \textit{learning} setting, where a set of solved motion planning problems is used to improve the efficiency of motion planning on different, yet similar problems. 
This setting is important in applications with rapidly changing environments such as in e-commerce, among others.
We investigate a general deep learning based approach, where a neural network is trained to map an image of the domain, the current robot state, and a goal robot state to the next robot state in the plan. 
We focus on the learning algorithm, and compare supervised learning methods with reinforcement learning (RL) algorithms. 
We first establish that supervised learning approaches are inferior in their accuracy due to insufficient data on the boundary of the obstacles, an issue that RL methods mitigate by actively exploring the domain.
We then propose a modification of the popular DDPG RL algorithm that is tailored to motion planning domains, by exploiting the known model in the problem and the set of solved plans in the data.
We show that our algorithm, dubbed DDPG-MP, significantly improves the accuracy of the learned motion planning policy. 
Finally, we show that given enough training data, our method can plan significantly faster on novel domains than off-the-shelf sampling based motion planners.
Results of our experiments are shown in \url{https://youtu.be/wHQ4Y4mBRb8}.

\end{abstract}

\IEEEpeerreviewmaketitle

\section{Introduction}
% \TJ{according to what you wrote below, I assume that this means that "intro" is fused with "related work" right?}\AT{We should have a dedicated related work section that discusses other MP+learning algorithms}
% 1. Motion planning is important. \\
Motion planning -- the problem of finding a collision free trajectory for a robot -- is a fundamental component in almost all robotic applications deployed today~\cite{lavalle2006planning,latombe2012robot}. Sampling based motion planners such as probabilistic roadmaps~\cite{kavraki1994probabilistic} and rapidly exploring random trees~\cite{latombe2012robot} have been studied extensively, can be guaranteed to converge to an optimal solution~\cite{karaman2011sampling}, and are common practice in various robotic domains and off-the-shelf software implementations~\cite{sucan2012the-open-motion-planning-library,moveit}. 
However, for domains where the environment can change rapidly, such as in e-commerce applications, industry 4.0~\cite{lasi2014industry}, or home robotics, it is desired to plan \emph{fast}, and the computational burden of sampling based planners can be limiting. 

% As robots are becoming more and more common in many industries such as manufacturing assembly lines, warehouse managements, farming and delivery, it is important that they will have the ability to move around independently without a human operator.
% This problem, of robotic motion planning describes algorithmic solutions which allow robots to autonomously plan a path from a source to a target position, which is a fundamental skill for almost all robotic platforms.

% 2. The learning setting in motion planning is important (+refs). \\
Consider, for example, planning pick-and-place motions for a robot in an e-commerce warehouse that handles various products. Building on the insight that the changes in the task will mostly be limited, e.g., just the product will change, while the robot and surrounding will stay the same, several recent studies have explored a \emph{learning} setting for motion planning, where data from previously solved motion plans is used to speed-up planning in new domains~\cite{berenson2012robot,jetchev2010trajectory,dey2012efficient,dragan2017learning}.

% In recent years companies try to design robots which are versatile enough to operate in the same space as humans doing similar tasks \todo{self driving reference or co-bots?}, navigating in hard terrain \todo{which ref??} or doing complex maneuvers \todo{boston dynamics??}.
% This requires that the robots have efficient and robust algorithms that could tackle various obstacles, environments and situations perceived by sometimes noisy sensors. 
% Therefore the robot must think of a complex plan sometimes with many future steps and execute them with a high level of precision.
% Learning can allow the robot to generalize past solutions for similar problems tackling completely new problems efficiently and accurately.

% 3. Deep learning has the potential to learn complex patterns in the data that are important for making decisions (ref alpha go, alexnet, protein folding, etc.). Here, we investigate whether deep learning can learn to map features from a motion planning domain to a suitable motion plan. (ref similar approaches for other planning problems such as TSP, Sokoban, etc.) \\
% Deep learning has the potential to learn complex patterns in the data that are important for making decisions \todo{(ref alpha go, alexnet, protein folding, etc.)}, and was previously used for planning \todo{TSP, Sokoban}.
% Here, we specifically investigate whether deep learning can learn to map features from a motion planning domain to a suitable motion plan. 

In recent years, deep learning has proven capable of learning complex patterns in data for various decision making domains such as computer vision, protein folding, and games~\cite{krizhevsky2012imagenet,alphafold,silver2016mastering}. Motivated by these successes, we focus here on approaches that we collectively term \emph{neural motion planners}~\cite{pfeiffer2017perception,bency2018towards,zhang2018auto,qureshi2018deeply,qureshi2018motion}, which use deep learning to approximate a motion planning computation. 
In a neural motion planner, a deep neural network is trained to map features of the domain (e.g., an image), the current robot state, and a goal robot state to the next robot state in the motion plan. By training on a set of motion planning domains, the network is hypothesized to learn the patterns which make for a successful motion plan, and, once trained, such a network can be used to quickly predict a motion plan in novel domains without running a heavy motion planning computation.

% 4. We first consider a supervised learning approach, where previous plans are simply imitated (ref some previous works that do this). We observe that for high dimensional domains with tight passages the accuracy of this approach is limited, which we attribute to insufficient data on the boundary of the obstacles. \\
% 5. We then consider an RL approach that learns through trial and error. We show that the data in this approach is better represented on the obstacle boundaries, leading to improved results. However, the accuracy is still limited. \\
In this work, we investigate the algorithmic aspects of training a neural motion planner to solve nontrivial motion planning problems.
We first consider a supervised learning approach, where previous plans are simply imitated~\cite{qureshi2018motion,groshev2018learning}. 
We observe that for high-dimensional domains that require high precision, the success of this approach is limited, which we attribute to insufficient data distribution on the boundary of the obstacles. We propose that reinforcement learning (RL; \cite{sutton1998reinforcement}) has the potential to overcome this problem, since the exploration process in RL will naturally drive the agent to investigate important areas in the domain, such as obstacle boundaries.

% We then consider an RL approach, based on a successful previous work Deep Deterministic Policy Gradient (DDPG) \todo{ref} that learns through trial-and-error and show that the data in this approach is better represented on the obstacle boundaries, leading to improved results. 
% However, this algorithm is not as accurate as required by many robotic motion planning situations as we show in experiments we conducted.

% 6. Our main contribution is an RL algorithm that is tailored to motion planning problems. Our algorithms does... We show that these modifications leads to significantly better accuracy. \\
While RL algorithms are known to require extensive computation, the motion planning problem presents several features that can be exploited to dramatically improve RL performance. Our main contribution is an RL algorithm that exploits
% is a modification to the DDPG algorithm that is tailored to 
two features of motion planning problems: the fact that a model of the dynamics is known, and the fact that we can collect demonstration data offline through sampling based planners, to reduce the variance in training, and to perform efficient exploration.
% Like many other model-based RL approaches \todo{ref some} our algorithm DDPG-MotionPlanner (DDPG-MP) first learns a world model.
% This model then allows to reduce some of the hardness during the training process of the agent. 

% Another important contribution of our model is smart exploration.
% In general RL approaches need to balance the need for exploration to discover new things, to exploitation which make the most of previously attained knowledge.
% In most algorithms the main source of exploration strategies relies on injecting randomness to the actions, thus uncovering states which the agent is not primed to visit.
% Another strategy which was used in the past is Hindsight Experience Replay \todo{cite} where agents attempt to reflect on past mistakes and redefine goals to formulate these mistakes as successes.
% This strategy is great for rapid exploration, however as we show in this work, when long sequences of specific actions are required neither of these approaches work.
% Instead we propose to incorporate expert demonstrations given by off-the-shelve motion planners into the RL agent's experience during \textbf{training}.
% We do so by harnessing the model for scoring these off-policy transitions, and as we show this allows the robot to achieve almost perfect accuracy even on domains with very challenging narrow passages.
% We show that these modifications leads to significantly better accuracy, and much faster training times.

We show that our method leads to significantly better accuracy, and much faster training times. In particular, we demonstrate predicting motion plans with almost perfect accuracy on a 4-dimensional robotic arm domain with very challenging narrow passages. 
With enough training time, our method learns to plan significantly faster than off-the-shelf sampling based planners. 
Our results suggest that with suitable training algorithms, deep learning can provide competitive results for learning in motion planning, opening the door to further investigations of network architectures and combinations of planning and learning.
% We show that these modifications leads to significantly better accuracy, and much faster training times.

% 7. We show that our method can plan significantly faster than off-the-shelf sampling based planners on a 4-dimensional robot arm domain. \\
% 8. Our results suggest that with suitable training algorithms, deep learning can provide competitive results for learning in motion planning, opening the door to further investigations of network architectures and combinations of planning and learning.

\begin{figure*}\label{fig:full-hard-trajectory}
\centering
\includegraphics[width=0.7\textwidth]{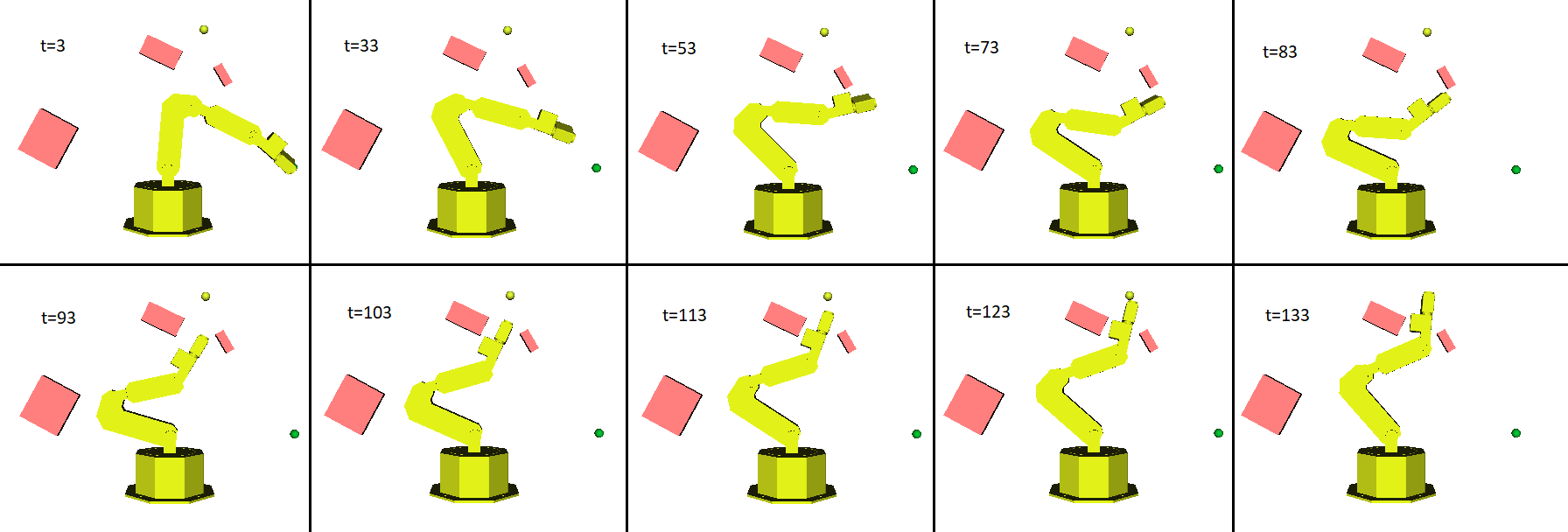}
\caption{DDPG-MP solves a narrow passage work-space. The figure shows several joint configurations along a successful plan executed by DDPG-MP for a narrow passage work-space (dubbed the \textit{hard} scenario in Section \ref{sec:results}). The episode starts at top-left image and ends at the bottom right. The green and yellow spheres mark the starting-pose and  end-pose of the end-effector, $t$ marks the time-step. Note the high precision required to navigate the arm through the two right obstacles without collision.}
\end{figure*}

\section{Background}
We start by describing several mathematical frameworks which will be useful when we present our algorithm in Section \ref{seq:model}, such as the motion planning problems, and several learning frameworks that could be used to solve it.

\subsection{Motion Planning}
We follow a standard motion planning formulation~\cite{lavalle2006planning}. Let $C$ denote the configuration space (joint space) of the robot, let $S$ denote the task space (position and orientation) of the end effector, and let $FK: C\rightarrow S$ be the robot's forward kinematics. Let $F \subset C$ denote the free space, and $\phi: C \rightarrow \{True, False \}$ a collision predicate that returns $True$ if $c\in F$. In the motion planning problem, the robot starts from a joint position $c_0\in C$, and is required to produce a trajectory plan $T:[0,1]\rightarrow C$ in the free space that ends with the end-effector at a goal pose $g\in S$:
\begin{equation}\label{eq:motion_planning_constraints}
    T(0)=c_0, \!\! \quad FK(T(1))=g, \!\!\quad \forall a\in[0,1]: \phi(T(a))=True.
\end{equation}
Popular motion planning algorithms rely on sampling the workspace, and effectively return a list of configurations $c_0,
\dots,c_T$ which make up the motion plan. A continuous plan is then derived by interpolation or smoothing~\cite{latombe2012robot}.
We denote by $W=\{ F, c_0, g\}$ a tuple that describes the \textit{workspace} -- the parameters of a motion planning problem for a given robotic domain.

\subsection{Reinforcement Learning}
In Reinforcement Learning (RL), an agent interacts with an environment $E$ by observing the current state and executing an action in discrete time steps. 
Once an action is executed the state changes (the change can be either deterministic or stochastic), the agent receives a reward and the process repeats until termination. 
This can be modeled as a Markov Decision Process (MDP~\cite{sutton1998reinforcement}) with a state space $S$, action space $A$, an initial state distribution $\rho(s_1)$, transition dynamics $p(s_{t+1} | s_t, a_t)$, and scalar reward function $r(s_t,a_t)$.

The goal of the agent is to interact with $E$ by selecting actions according to some policy $\pi(a|s)$ (possibly stochastic) which maximizes the $\gamma$-discounted future rewards defined for time step $t$ as $R_t = \mathbb{E} \left[ \sum_{i=t}^{\infty}{\gamma^{i-t}r(s_i,a_i)}\right]$.
Let $Q^\pi(s,a)$ be the expected discounted future reward obtained by following policy $\pi$, after executing action $a$ from state $s$: $Q^\pi(s,a)= \mathbb{E}\left[R_t | s_t=a, a_t=a, \pi\right]$. 
An important relation between current and future $Q^\pi$ is the Bellman equation:
\begin{align} \label{eq:bellam_general}
\begin{split}
Q^\pi &(s_t,a_t) = \\
&\mathbb{E}_{s_{t+1} \sim p} \left[r(s_t,a_t) + \gamma \mathbb{E}_{a_{t+1} \sim \pi(s_{t+1})}\left[Q^\pi (s_{t+1}, a_{t+1})\right]\right].
\end{split}
\raisetag{2.5em}
\end{align}
Let $Q^*(s,a)$ be the maximal expected discounted reward obtained by any strategy, after executing action $a$ from state $s$: $Q^*(s,a)=\max_{\pi}{\mathbb{E}[R_t | s_t=a, a_t=a, \pi]}$. 
For the optimal policy $\pi^*$ we have $Q^{\pi^*}(s,a) = Q^*(s,a)$ for all states and actions.
The goal of an RL algorithm is to find such a policy.

\subsection{Actor-Critic Algorithm: DDPG}
Actor-critic is a popular RL approach~\cite{sutton1998reinforcement}.
The actor $\pi(s|a)$ models a policy, while the critic $\hat{Q}(s,a)$ approximates the state-action value of the actor.
The learning process interleaves improving the critic (for example, by using temporal difference learning~\cite{sutton1998reinforcement}), with updating the policy towards actions with higher state-action values. 
% as determined by the critic.
% One approach to solve RL problems is the actor-critic approach. 
% In this approach an actor $\pi$ describes the policy, and the critic approximates the state-action value $\hat{Q^\pi}$.
% To learning process interleaves improving the critic approximation as described by equation \ref{eq:bellam_general}, with updating the policy towards actions with higher (approximated) state-action values $\hat{Q^\pi}$.

Deep Deterministic Policy Gradient (DDPG~\cite{lillicrap2015continuous}) is an actor-critic RL algorithm designed for continuous state and action spaces, where $\pi$ and $\hat{Q}$ are modeled using neural networks, and $\pi$ is a deterministic policy with weights $\theta$. 
DDPG is an off-policy algorithm: it learns by observing actions taken by a stochastic exploration policy $\beta$, and storing the observed transitions in a \textit{replay buffer}~\cite{mnih2015human}.
DDPG alternates between learning $\hat{Q}$ and learning $\pi$, fixing one while training the other. 
Specifically, the update of $\pi$ ascends in the direction which maximizes $\hat{Q}$, motivated by the policy gradient theorem:
% (while considering $\hat{Q}$ fixed) 
\begin{equation}\label{eq:ddpg-actor-update}
    \nabla_\theta J \approx \mathbb{E}_{s \sim \beta} \left[ \nabla_\theta \hat{Q}(s,a) \right],
\end{equation}
where $J$ denotes the expected discounted return~\cite{lillicrap2015continuous}.

\subsection{Hindsight Experience Replay}\label{sec:her}
In NMP, the learned policy is \emph{goal-conditioned}, i.e., takes in a goal as a parameter. The recent 
Hindsight Experience Replay (HER)~\cite{andrychowicz2017hindsight}, is an extension of off-policy RL methods such as DDPG to goal-conditioned policies. 
The idea is that, upon failing an episode (not reaching the goal), the agent adds to the replay-buffer an episode comprised of the same states and actions seen in the failed episode, but with the original  goal replaced with the last encountered step -- making the episode a success in hindsight.
By making success more frequent, the agent is trained by an implicit curriculum, that eventually learns to cover the goal space.
% little-by-little in hopes of covering the goal space fully as training progresses. 
Indeed, HER has led to state-of-the-art results in goal-conditioned RL~\cite{andrychowicz2017hindsight,srinivas2018universal,levy2018hierarchical}.   

% In the recent work on universal planning networks, HER was used to train an NMP as a state-of-the-art exploration strategy, also in the motion-planning context \cite{srinivas2018universal}.

\subsection{Imitation Learning}
In imitation learning~\cite{schaal2003computational}, a decision making policy is learned from observing expert behavior. Following the RL notation, given a data set of $N$ expert demonstrations $\{ s^{exp}_i, a^{exp}_i\}_{i=1}^N$, the problem is to find a policy $\pi = P(a|s)$ that imitates the expert behavior. A popular imitation learning method is behavioral cloning~\cite{pomerleau1989alvinn}, where $\pi$ in learned by maximizing the data log likelihood $\sum_{i=1}^{N}\log \pi(a^{exp}_i|s^{exp}_i)$.

DAgger~\cite{ross2011reduction} is an imitation learning algorithm that corrects the \textit{distribution mismatch problem}, where the distribution of states in the learning corpus may be different than the distribution of states encountered by the learned policy. 
In DAgger, after training a policy using behavioral cloning, the policy is executed, and additional expert data is first collected for all states visited, and then added to the agent's data-set. Afterwards, a new policy is trained on the updated data-set and the process repeats to convergence.
% To overcome the issue, DAgger tries to match these distributions during the course of training by starting with an initial small data-set of trajectories and after every training epoch collect more trajectories, let the expert predict the correct action from each encountered state, and finally add these new transitions to the data-set.

\section{Problem Formulation}
We consider the problem of learning for motion planning. In this setting, we assume that motion planning workspaces are generated from some (unknown) distribution $P(W)$. We are given a training data set of $N$ workspaces $\{ W_i \}_{i=1}^N$ drawn from $P$, and the goal is to use this data to plan faster in an unseen test workspace $W_{test}$, also drawn from $P$. 

We focus on an approach termed neural motion planning (NMP)~\cite{qureshi2018motion}. In NMP, a workspace $W$ is described by a set of features, or context vector, $w$. The goal in NMP is to train a neural network policy $\pi$ that maps $w$ and the current configuration of the robot $c_t$ to the next step in a motion plan $c_{t+1}$. Once trained, such a network can produce a motion plan by unrolling the predictions to produce a list of states $c_0,\dots,c_T$. We say that the motion plan is successful if it satisfies the motion planning problem conditions, i.e., $FK(c_T)=g$, and $\phi(c_t)=True \quad \forall 0\leq t \leq T$.\footnote{In practice, we can only verify that $\|FK(c_T)-g\| < \epsilon$, for a small $\epsilon$.}
In the remainder of this work, we investigate how to best train the neural network in NMP. 
% \AT{This could be the place to discuss approximately reaching the goal condition}

\section{Imitation Learning vs. RL}\label{sec:IL-vs-RL}
A straightforward method for training an NMP is using imitation learning. In this approach, a standard motion planner (e.g., RRT$^*$) is used to solve all the training work spaces. These motion plans are used to train a neural network using, e.g., behavioral cloning. This approach has been shown to work well for discrete planning problems~\cite{groshev2018learning,dai2017learning}, and recently has been investigated in the context of NMP~\cite{qureshi2018motion,bency2018towards,zhang2018auto}.

\begin{figure}
    \centering
    \includegraphics[width=0.65\columnwidth]{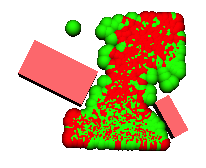}
    \caption{Data Distribution on Edges of Obstacles. Spheres indicate visited end-effector poses seen during training. Green spheres were generated by an RL agent (DDPG) while red spheres are imitation learning targets (i.e., produced by the motion planner). Note that near obstacle boundaries, green sphere are more dominant, indicating the sparsity of data that imitation learning agents have on the edges of obstacles.}
    % , as seen by lack of red spheres on the exterior of the spheres volume which is dominated by green spheres. }
    \label{fig:data_distribution}
\end{figure}

In our investigation, we found that for domains that require high precision in the motion plan, such as a robot moving through tight passages, the performance of imitation learning is severely limited (see explicit results in Section~\ref{sec:results}). We attribute this finding to the distribution of the training data, as depicted in figure~\ref{fig:data_distribution} for the end effector of a robot arm moving in a narrow passage. In NMP, it is fundamental that the neural network learns not to hit obstacles. However, since the expert produces perfect motion plans, there is no information about hitting an obstacle in the training data. Note in Figure~\ref{fig:data_distribution} that the data distribution for imitation learning does not cover the obstacle boundaries, thus there is no reason to expect that a network trained with this data will learn not to hit obstacles. This is an instance of the 
% compounding errors 
distribution mismatch
problem described by~\citet{ross2011reduction}. While in principle the DAgger algorithm~\cite{ross2011reduction} could mitigate this issue, it is too costly to compute in practice as it requires running motion planning on every sample in the data. One can instead acquire expert data only for states that resulted in a collision. However, as we report in the supplementary material (Section \ref{sec:trainig-IL-agents}), this approach did not lead to a significant improvement, which we believe is due to theses samples not being well-balanced with the rest of the data.

We propose that RL is more suitable for training NMPs. In RL, the agent starts out with a policy that does not solve the task, and through trial and error it gradually collects data on how not to hit obstacles, thereby improving its policy. Indeed, as depicted in Figure~\ref{fig:data_distribution}, the data distribution in RL has much more presence near the obstacle boundaries -- the critical decision points for motion planning. Therefore, in this work we pursue an RL approach to NMP, as we describe next.

\section{RL for Motion Planning}\label{seq:model}
In this section we first describe an RL formulation for NMP, and then propose a specialized RL algorithm for this domain.

We consider an episodic RL setting~\cite{sutton1998reinforcement}, where at each episode a workspace $W$ is randomly selected from the training set, and the task is to plan a successful motion plan in it. At time step $t$ of an episode, the robot configuration is $c_t$ and the context vector for $W$ is $w$. We define the state $s_t$ to be the tuple $\{w, c_t\}$, and the action to be the difference in the robot configuration:\footnote{Our work can be extended to kinodynamic motion planning by including the robot's dynamics. We defer this to future work.}
\begin{equation}\label{eq:dynamics}
    c_{t+1} = c_t + a_t.
\end{equation}
Note that this implies deterministic dynamics, since the context vector $w$ is assumed to remain constant. The next state is:
\begin{equation}\label{eq:state_dynamics}
    s_{t+1} = \{w, c_t + a_t\} \doteq f(s_t,a_t). 
\end{equation}
% \begin{equation}\label{eq:state_dynamics}
% \begin{split}
%     s_{t+1} &= \{w, c_t + a_t, FK(c_t + a_t)\} \\
%     &\approx \{w, c_t + a_t, c_t + J^\top a_t)\} \doteq f(s_t,a_t),
% \end{split}
% \end{equation}
% where $J$ is the forward kinematics Jacobian, and we assume that movements are small.

Upon hitting an obstacle, or after reaching the goal, the episode terminates. We define a reward such that maximizing it produces a correct solution to the motion planning problem, by encouraging movement toward the goal and not hitting obstacles. Note that any transition $\{c_t, c_{t+1}\}$ can belong to one of three classes: free space movement, denoted as $T_{free}$; collision, denoted as $T_{col}$; and reaching the goal, denoted as $T_{goal}$. The transition reward is:
% Every transition $(c,\delta_c,w)$ belongs to one of three step classes depending on the motion it describes: $F$, $\bot$ and $\top$ for free space movement, collision and close-to-goal respectively.

% The reward in each time step is comprised of two parts $r(c,\delta_c,w)=r_l(c,\delta_c,w)+r_q(c,\delta_c,w)$.
% $r_l$ is the reward portion based on the transition class:
\begin{equation}\label{eq:reward}
    r^T_t=
    \begin{cases}
    -\epsilon, & (c_t,c_{t+1})\in T_{free}, \\
    1, & (c_t,c_{t+1})\in T_{goal}, \\
    -1, & (c_t,c_{t+1})\in T_{col}.
    \end{cases}
\end{equation}
The constant $\epsilon>0$ needs to be small enough such that the robot prefers moving in free space than colliding; we empirically set it as $0.01$. 

Some motion planning problems may include additional rewards. For example, in our experiments we wanted to encourage the robot not waste effort by trying to move a joint beyond its limits. To model this, we use a reward of $r^D_t$ (domain specific reward), which in our case is the norm of the wasted movement (full details in section \ref{sec:experiments-and-parameters}). Finally, the full reward of the system is $r_t = r^T_t + r^D_t$ .

\subsection{The DDPG-MP Algorithm}
We next present Deep Deterministic Policy Gradient for Motion Planning (DDPG-MP) -- a deep RL algorithm tailored for training an NMP. To motivate the algorithm, we observe that in the context of RL, the motion planning setting admits the following unique features:
\begin{enumerate}
    \item The dynamics (Eq. \eqref{eq:dynamics}) and reward (Eq. \eqref{eq:reward}) are known.
    \item Similar to imitation learning, we can obtain expert demonstrations for the work spaces in our training data.
\end{enumerate}
Since our problem is deterministic, we build on the DDPG algorithm~\cite{lillicrap2015continuous}, a strong and popular deep RL algorithm for continuous control. We add to DDPG two modifications that exploit the motion planning features described above: we propose a model-based actor update that reduces variance, and use the expert demonstrations to perform efficient exploration. We next describe each idea in detail (for the DDPG-MP pseudo-code see Section \ref{sec:algorithm} in the supplementary material).

\subsection{Model-Based Actor Update}\label{sec:ddpgmp-model-based-actor-update}
% \TJ{added powers of $\gamma$ in the recursive equations below}
When training the actor network, the model is updated in the direction which maximizes $Q^\pi$. However, $Q^\pi$ is not known, and we only have access to an approximation $\hat{Q}^\pi$ learned by the critic, resulting in a high estimation errors in the policy gradient (due to variance and bias) of the actor. A key observation in our work is that we can reduce the variance in the actor update by using the known dynamics model. Note that in a deterministic domain with a deterministic policy, by definition (cf. Eq.~\eqref{eq:bellam_general}) we have that for any $k\in 0,1,\dots$:
\begin{equation*}
\begin{split}
    Q^\pi (s_t,a_t) = &r(s_t,a_t) +\dots +\gamma^{k-1} r(s_{t+k},\pi(s_{t+k}))\\
    &+\gamma^k Q^\pi (s_{t+k+1}, \pi(s_{t+k+1})).
\end{split}
\end{equation*}
Thus, if we know the reward function and transition function, we can estimate $Q^\pi$ as $r(s_t,a_t) +\dots +\gamma^{k-1} r(s_{t+k},\pi(s_{t+k}))+\gamma^k \hat{Q}^\pi (s_{t+k+1}, \pi(s_{t+k+1}))$, and we expect that as $k$ grows, the error in this approximation will reduce~\cite{heess2015learning} (for $k\to \infty$ the error is zero). In our experiments, we found that $k=1$ is enough to significantly improve the actor update, and in the following we focus on this case. Extending our result to $k>1$ is straightforward. 

Following the above derivation, keeping in mind that transitioning to a goal or obstacle terminates the episode, and using the specific form of the dynamics in our problem \eqref{eq:state_dynamics}, we obtain the following actor update: 
% for non terminal transition i.e $(c_t,c_{t+1})\in T_{free}$:
% \TJ{we can use an indicator [see commented out equation below. But this is way too long for this double column format}
\begin{equation}\label{eq:ddpg-mp-actor-update}
\begin{split}
    \nabla_\theta J = &
    \mathbb{E}_{s \sim \beta} \big[ 
            \nabla_\theta \big(
                r(s_t,\pi(s_t)) \\
                &+ \gamma\mathbb{I}_{(c_t,c_{t+1})\in T_{free}} Q(f(s_{t},\pi(s_t)),\pi(f(s_{t},\pi(s_t))))
            \big)
        \big]
\end{split}
\raisetag{3em}
\end{equation}
% \begin{equation}\label{eq:ddpg-mp-actor-update}
%     \nabla_\theta J = 
%     \mathbb{E}_{s \sim \beta} \left[ 
%             \nabla_\theta \left(
%             \begin{split}
%                 &r(s_t,\pi(s_t)) \\
%                 &+ \gamma Q(f(s_{t},\pi(s_t)),\pi(f(s_{t},\pi(s_t))))
%             \end{split}
%             \right)
%         \right].
% \end{equation}
% Otherwise, 
% \begin{equation}\label{eq:ddpg-mp-actor-update-terminal}
%     \nabla_\theta J = 
%     \mathbb{E}_{s \sim \beta} \left[ 
%             \nabla_\theta \left(
%                 r(s_t,\pi(s_t))
%             \right)
%         \right].
% \end{equation}

Note, however, that both the indicator function in \eqref{eq:ddpg-mp-actor-update} and our reward function \eqref{eq:reward} are non differentiable, and their step-function structure means that the reward gradients in \eqref{eq:ddpg-mp-actor-update} will not be informative. 
% Moreover, we need a mechanism to decide which equation \ref{eq:ddpg-mp-actor-update} or \ref{eq:ddpg-mp-actor-update-terminal} to use.

We therefore propose to use a smoothed reward function $\widetilde{r}$ and a smooth approximation $\widetilde{p}$ to $\mathbb{I}_{(c_t,c_{t+1})\in T_{free}}$ instead. While one can derive explicit smoothing functions for these, in our work we opted for a general approach using supervised learning. Specifically, we represent $\widetilde{r}(w,c_t,c_{t+1})$ and $\widetilde{p}(w,c_t,c_{t+1})$ as neural networks, and train them using random samples of robot configurations. Technical implementation details regarding these models are given in Section \ref{sec:reward-models} of the supplementary material.\footnote{There are many possible choices for reward smoothing. We chose to model it as a neural network since it was easy to implement in out setup.}

\subsection{Targeted Exploration via Expert Demonstrations Scoring}\label{sec:targeted-exploration}
Another disadvantage of DDPG is sample inefficiency that occurs because of uninformed exploration.
DDPG's exploration policy selects a random action in every step with low probability to discover novel moves. This exploration policy makes succeeding in narrow passages highly unlikely, since these often require a very specific action sequence with no guiding reward signal along the way. Since the probability to randomly generate such a sequence of actions is extremely low, the agent can never reach the other side of the passage to collect the reward. Worse, randomly exploring this narrow corridor leads in practice to many collisions, making the agent avoid entering the passage in the first place.
% as it goes back-and-forth in a local minima.

In this work we propose to overcome the issue by providing the agent with motion planner demonstrations for failed work spaces. 
This is especially important in tight passages since instead of failing over-and-over, after the first failure the agent gets a useful signal that it can use to learn appropriate actions.
Note that this strategy implicitly generates a curriculum: easier failure cases resolve earlier during training and the agent stops receiving demonstrations for those, while harder cases are still supplied with demonstrations. At the end of training we expect to succeed in most attempts and thus not use this exploration strategy, which is inline with other exploration strategies where the noise is being reduced over time, e.g using an $\epsilon$-greedy approach with a decreasing $\epsilon$ over time~\cite{sutton1998reinforcement}.

Let $\tau_W = (c^\tau_0, a_0, c^\tau_1, a_1 ... a_{N-1}, c^\tau_N)$ be a trajectory produced by a motion planner, which is a solution to a failed workspace query $W=\{ F, c_0, g\}$, i.e $c^\tau_0=c_0$ and $FK(c^\tau_N)=g$.
In order to incorporate the information of $\tau_W$ to the replay buffer, we only need to assign a reward to every transition, to obtain $(c^\tau_i, a_i, r_i, c^\tau_{i+1})$.
Because $\tau_W$ is known to be a valid trajectory, this process is straightforward -- by applying Eq. \eqref{eq:reward}, to obtain $r_i$.
% We note that if $r^D_t$ is not negligible, one can use $\widetilde{r}$ as an approximation instead.\AT{previous sentence is not clear to me}
This strategy can be seen as a `smart' form of exploration, where instead of trying random actions, we are guided by the motion planner to try actions that result in success. We have found that this allows us to reduce the magnitude of the random exploration required for DDPG to work, thereby obtaining a much more stable algorithm.
% allows DDPG-MP to use much smaller random exploration parameters since novel actions can be demonstrated by the expert.

\begin{figure*}
\centering
\begin{subfigure}{0.27\textwidth}
\includegraphics[width=\textwidth]{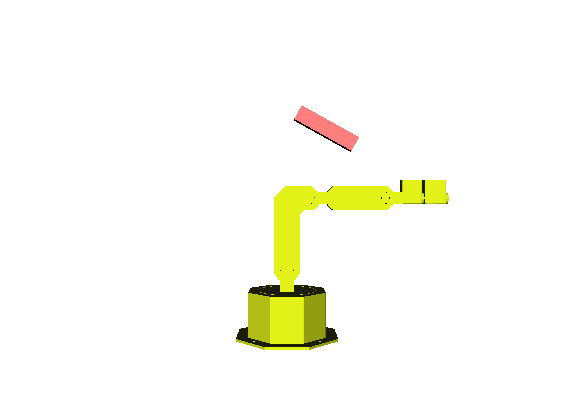}
\caption{\textit{Simple} scenario - single obstacle.}
\label{fig:simple-scenario}
\end{subfigure}~
\begin{subfigure}{0.27\textwidth}
\includegraphics[width=\textwidth]{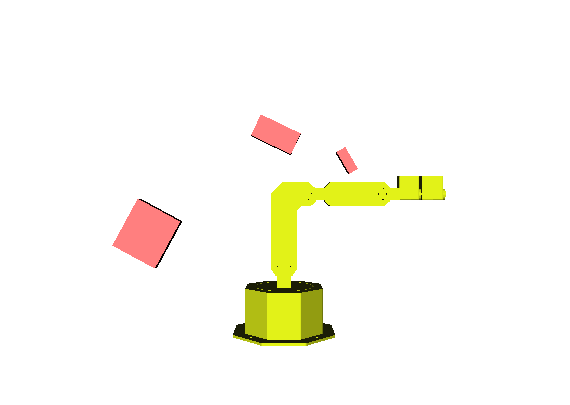}
\caption{\textit{Hard} scenario - three obstacles. The top two obstacles create a
narrow passage. }
\label{fig:hard-scenario}
\end{subfigure}~
\begin{subfigure}{0.27\textwidth}
\includegraphics[width=\textwidth]{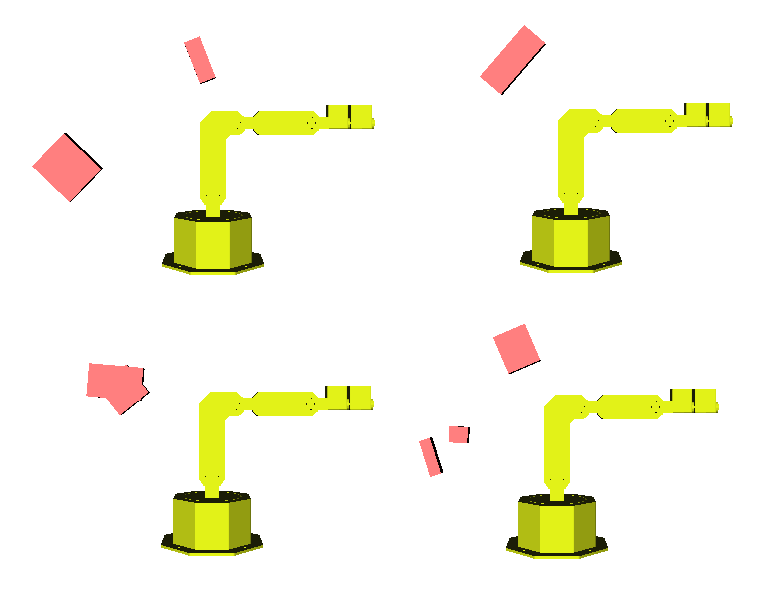}
\caption{\textit{Vision} scenario - 1000 obstacle configurations selected randomly.}
\label{fig:vision-scenario}
\end{subfigure}
\caption{Scenarios - the \textit{simple} (left), \textit{hard} (middle) and four random samples from the \textit{vision} (right) scenarios used in our experiments.}
\label{fig:scenarios}
\end{figure*}

\section{Related Work}
%NMP

The idea of learning from experience to plan faster in motion planning has been studied under various approaches, such as using a library of past trajectories~\cite{branicky2008path,jetchev2010trajectory,berenson2012robot,dey2012efficient}, learning the sampling procedure~\cite{martin2007offline,ichter2018robot,kuo2018deep}, learning a latent representation of obstacles~\cite{ichter2018robot}, and learning to select goals~\cite{dragan2017learning}.

The idea of using a neural network for motion planning (termed here as NMP) dates back to the previous decade~\cite{glasius1995neural,yang2000efficient}. Following recent advances in deep learning~\cite{krizhevsky2012imagenet}, interest in these methods has rekindled~\cite{pfeiffer2017perception,bency2018towards,zhang2018auto,qureshi2018deeply,qureshi2018motion}. The potential advantages of NMP are the possibility of generalizing to different domains, working directly on raw perceptual signals such as images, and a low memory footprint: once trained, the NMP is a compact prediction model that does not require storing a large library of trajectories. While most previous work on NMP concerned the neural network architecture~\cite[e.g.,][]{lebedev2005dynamic,qureshi2018motion,zhang2018auto,bency2018towards}, in this work we focus on a parallel investigation -- the training algorithm for NMP. We believe that with better architectures, our results can be further improved.

RL for continuous control has been explored extensively~\cite{schulman2015trust,lillicrap2015continuous}. Our method builds on two main ideas for improving RL: using expert demonstrations, and a model-based update for the policy gradient. While similar ideas have been explored in the literature~\cite[e.g.,][]{peng2018deepmimic,pfeiffer2018reinforced,heess2015learning}, our formulation is tailored for the NMP setting, and is, to the best of our knowledge, novel. 

Using neural networks for planning in discrete tasks has been explored for navigation~\cite{tamar2016value}, task planning and combinatorial optimization~\cite{groshev2018learning,dai2017learning} and games~\cite{silver2016mastering}. Interestingly, the seminal AlphaGo program~\cite{silver2016mastering} exploited expert demonstrations and the known structure of the game to improve RL. Our work exploits similar properties of continuous motion planning domains, therefore requiring different methods.

\section{Results}\label{sec:results}

In this section we evaluate the DDPG-MP algorithm on various NMP domains, and show that it obtains state-of-the-art results in training an NMP model. In addition, we investigate the components that contribute to DDPG-MP's success using an ablation study. Finally, we show that using DDPG-MP we can significantly improve upon the planning times of conventional motion planners.
% causes for these gains in an ablation study, designing experiments that test each of our contributions individually. 
% We show that  currently state-of-the-art approach of using DDPG+HER obtaining large success rate gains as demonstrated by figures \ref{fig:success-plot-simple} and \ref{fig:success-plot-hard} and tables \ref{tbl:hard-scenatio-success-rates} and \ref{tbl:vision-scenatio-success-rates}. 
% Specifically, DDPG-MP is the model with highest success rate in every experiment conducted, and on an extremely challenging scenario, previous approaches only reached $0.31$ success rate, while DDPG-MP reached a near perfect success rate of $0.97$.
% We also investigate the causes for these gains in an ablation study, designing experiments that test each of our contributions individually. 
% We also motivate the use of RL algorithms in the first place, by showing that IL does not work as expected. 
We investigate the following questions:
% \TJ{general arch, we compare the algo to SOTA (ddpg + her) we are way better, we want to understand why, we dessigned experiment to understand how each component adds value.}
% In this section we demonstrate experiments which validate the claims from previous sections regarding both the contributions of DDPG-MP, and motivating the use of an RL algorithm in the first place. 
% Specifically, we will consider the following:
\begin{enumerate}
    \item Is RL more suitable for NMP than imitation learning?
    \item Can DDPG-MP obtain state-of-the-art results for NMP?
    \item Can NMP models trained with DDPG-MP generalize to unseen work spaces with a high success rate?
    \item Can we use NMP models to plan faster than sampling based motion planners? 
\end{enumerate}

% \TJ{add references to works that do!}
We next describe the experimental settings we used.

\textbf{NMP Architecture:} As outlined above, in this work we wish to disentangle the questions of NMP architectures from the NMP training algorithm. Therefore, we chose popular neural network architectures that were the same for all methods we tested. Specifically, for domains without image inputs we used fully connected neural networks, which are popular for continuous control~\cite{duan2016benchmarking}. For domains with image inputs we used convolutional layers followed by fully connected layers~\cite{mnih2015human,lillicrap2015continuous}. 
% and for domains and focus our effort in showing that it is our algorithmic contributions that provide the reported gains. The only exception to that is when we process an image input, and for that we copy a shallow version of DQN ~\cite{mnih2013playing}. 
We provide full details of the network architecture in Section \ref{sec:experiments-and-parameters} of the supplementary material.

\textbf{Simulation environment:} In our choice of a domain, we opted for an environment that can present challenging planning problems, yet be simple enough visually to not require complex perception efforts. We chose a 4DoF WidowX robotic arm~\cite{WidowX}, with movement restricted the XZ plane, and box-shaped obstacles in various sizes, positions, and orientations. Restricting movement to the plane allows us to capture the domain using a simple 2-dimensional image of the obstacles, which is natural for processing with convolutional neural networks, as discussed above. By varying the positioning of the obstacles, we can generate planning problems with various difficulty levels. We note that this setup is significantly more challenging than the point robots and car-like robots explored in previous NMP work~\cite[e.g.,][]{qureshi2018motion}.
% We modeled our experiments in the robotics simulation environment OpenRAVE~\cite{diankov2008openrave}, which we found fast enough to generate our training data.  and when we specify a use of a motion planner, it is OpenRAVE off-the-shelve default planner\AT{What is the default planner?}\TJ{I don't know, I found mentions online that say it is RRT based. I can't find it in the docs. Will keep looking}\AT{OK, will be good to have a name. I'm guessing it is bidirectional RRT, but give it a check.}.
% We simulate a 4DoF WidowX robotic arm~\cite{WidowX}, with movement restricted the XZ plane. 
% This allows us to employ standard conv-nets for the image processing, and focus in this work on the question presented above, instead of trying to also tackle the problem of modeling a 3D world. 
% On the other hand, this setup with a 4DoF arm, is still a more challenging problem than a point robot or other 2/3 DoF explored previously \todo{ref}. 

We used Tensorflow~\cite{tensorflow2015-whitepaper} for training the neural networks. 
% , which is a python deep-learning with auto-differentiation package. 
Our code is available at \url{https://github.com/tomjur/ModelBasedDDPG}.

% \AT{It's important, but needs to be polished. You can put in a paragraph titled evaluation metric. One thing that people will ask is how long are our trajectories compared to the motion planner. If you can measure that then great, if not leave for rebuttal.}
\textbf{Evaluation metric:} Evaluating the performance of agents trained with RL is a delicate matter~\cite{henderson2018deep}, and in particular, how to choose the `best' agent during training requires some validation metric. To address this, 
% In order to calculate the final success scores for models 
for each training run, the model iteration with highest test success-rate is taken and tested again on  $1000$ new validation work-spaces.
This asserts that the reported scores are not just a lucky sample, but accurately capture the success rate of the model (the validation success-rates are mostly lower than the best test success-rates).
% Also, the final success scores of the models are taken on $1000$ new validation work-spaces, not the best test scores obtained during training (the validation scores are usually lower).

% \AT{Explain why we are using a 4D domain (I think we can explain by saying that we wanted to use standard conv nets for the image, so didn't use higher dim, while this is still more challenging than 2/3dim which were explored in [refs]).}

\textbf{Scenarios:} Within the simulation described above, we conducted experiments on three scenarios, which we term \textit{simple}, \textit{hard} and \textit{vision}, as depicted in Figure~\ref{fig:scenarios}.
% \AT{Relate to the definition of $W$ - in simple,hard, $F$ is the same, and start and goal are selected randomly, and $w$ is.... In vision $F$ changes, and $w$ is an image.}
The \textit{simple} scenario (Figure~\ref{fig:simple-scenario}) contains just a single obstacle. 
% We use it to demonstrate specific advantages of our algorithm that would have been harder to analyze on the other two scenarios. 
The \textit{hard} scenario (Figure \ref{fig:hard-scenario}) was manually designed with a challenging obstacle configuration:
% , where all parts of DDPG-MP are required in order to reach a solution with near perfect success rate.
three obstacles, such that the two top obstacles create a narrow passage where a sequence of very specific movements is required in order to cross.
% Meanwhile, the bottom left obstacle restricts the movement of the robot especially when entering the narrow area from below.
For the NMP problem, in both \textit{simple} and \textit{hard} scenarios the obstacle configuration is fixed, and the only variation is in the starting and goal positions. Formally, $W=(F,c_0,g)$ contains a fixed $F$, and $c_0$ and $g$ are sampled from a distribution of feasible trajectories. The context vector $w$ is simply $g$.

The \textit{vision} scenario, shown in Figure \ref{fig:vision-scenario}, is used to investigate generalization to unseen obstacle configurations. We generated a data set of $1000$ random obstacle configurations, partitioned to $80$\% train and $20$\% test. We report success rates on the test set.
% The training of both the supervised models and the RL agents were executed only on the training set, and the work space success rates was measured on the $200$ unseen test set.
For this scenario all three parts of $W=(F,c_0,g)$ are randomly sampled, and the context vector in this case contains both an image $I_F$ of the obstacle configuration, and the goal pose $g$.

We note that when sampling random work spaces (either obstacles or starting positions and goals), trivial problems, such as problems that do not require crossing obstacles, or where the goal is very close to the starting point, are frequent. To focus on interesting planning problems, we used rejection sampling to bias our testing to problems that require crossing from one side of an obstacle to the other. We note that in the \textit{hard} and \textit{vision} scenarios, the arm often needs to cross more than a single obstacle, making the task even more challenging. 
The difficulty of our domain is best appreciated in the accompanying video \url{https://youtu.be/wHQ4Y4mBRb8}.
Full details of the sampling process are given in Section \ref{sec:experiments-and-parameters}.
We next present our results.
% Meaning that there exists at least one obstacle center $o$ such that $FK(c_0)$ and $g$ satisfy: $\|o-FK(c_0)\|_2 \le \|g-FK(c_0)\|_2$ and $\|o-g\|_2 \le \|g-FK(c_0)\|_2$
% To differentiate between obstacle configurations in the \textit{vision} scenario the context vector $w$ contains both an image of the obstacle configuration, and the goal pose $g$. For the other two scenarios, it is not required and $w$ only contains the goal pose. 
% For full details see section \ref{sec:experiments-and-parameters}.

% \AT{Discuss NN architectures. We can say that we chose standard architectures in order to focus the investigation on the algorithmic components. 3-layer FC are snatdard in continuous control, and the convnets follow the DQN architecture.}

% In the next subsection we attempt to give answers to the questions presented above. Section \ref{sec:IL-performace} starts with the fundamental decision of using RL over IL. The next two sections \ref{sec:actor-stabilization} and \ref{sec:harnessing-demonstrations} measure the contribution of each of our two improvements. 
% Finally we show that our model generalizes to unseen obstacle configurations in section \ref{sec:vision-generalization} and that the planning time is short in section \ref{sec:planning-times}.

\begin{figure}
\centering
\includegraphics[width=0.4\textwidth]{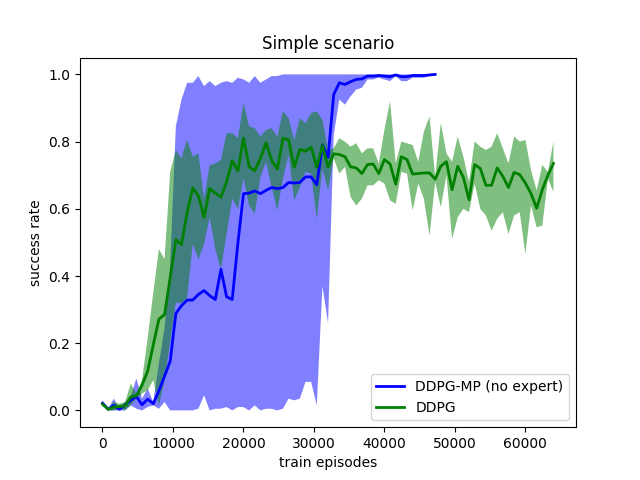}
\caption{\textit{Simple} scenario success rate: visual comparison between DDPG in green, to DDPG-MP (no expert) in blue. Each curve shows the average test success rate of 3 runs, with the upper and lower bounds the maximal and minimal values respectively. Each data-point is the count of successful work-spaces out of $200$ test work-spaces shown to the model during testing. }
\label{fig:success-plot-simple}
\end{figure}

\begin{figure}
\centering
\includegraphics[width=0.4\textwidth]{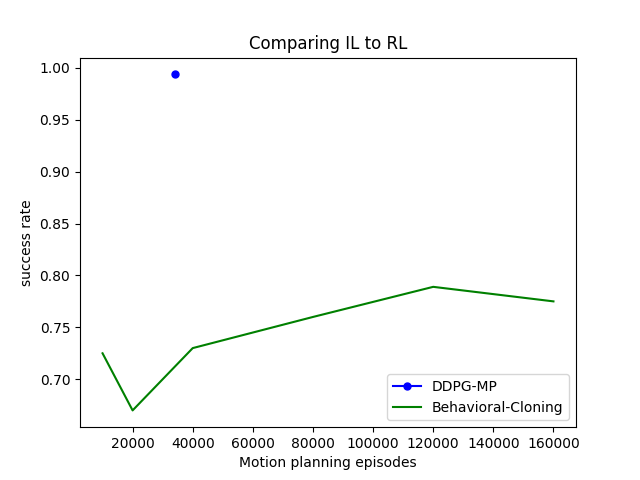}
\caption{Comparison between IL method to our method on the \textit{simple} workspace. }
\label{fig:il-vs-rl-plot}
\end{figure}

\subsection{Imitation Learning vs. RL}\label{sec:IL-performace}
As mentioned above, we have found that imitation learning does not provide accurate enough training for NMP. We demonstrate this here for the \textit{simple} scenario, while similar results were obtained for other scenarios.
% Before investigating RL solutions, we considered IL as the default training method.
% We wanted to see if an IL agent can reach a perfect success rate in the \textit{simple} scenario.
We collected up to 160K expert demonstrations and trained imitation learning agents using behavioral cloning and DAgger. The results are presented in Figure \ref{fig:il-vs-rl-plot}, for various sizes of the training data set.
The best behavioral-cloning and DAgger models reached a success rate of only $0.78$ and $0.8$ respectively. We also observed that the success-rate gains were diminishing, meaning that there is little contribution in adding more data.
RL, on the other hand, was able to obtain a near-perfect NMP with a fraction of the expert demonstrations. This is explained in Figure \ref{fig:data_distribution} by observing the data distribution of RL and imitation learning, and noting that for RL the data is better distributed in the important areas -- near obstacles.
% This observation motivated us to look for alternative training methods, and thus we turned to RL.
% Since motion planning takes time, collecting 160K motion plans is not desirable and we expected the IL model to have higher success rates with much less demonstrations. Indeed, as we see later, RL agents are able to make better use of the data and our algorithm DDPG-MP solves the \textit{simple} scenario in only 33K episodes.
% The full results for both behavioral-cloning and DAgger can be found in the supplementary material section \ref{sec:trainig-IL-agents}.
% \AT{Explain that in terms of number of expert trajectories required (the main time bottleneck), RL is much more efficient. In addition, it seems that IL plateaus on bad accuracy.}

\subsection{RL for NMP}
\begin{figure*}[h]
\centering
\begin{subfigure}[t]{0.35\textwidth}
\includegraphics[width=\textwidth]{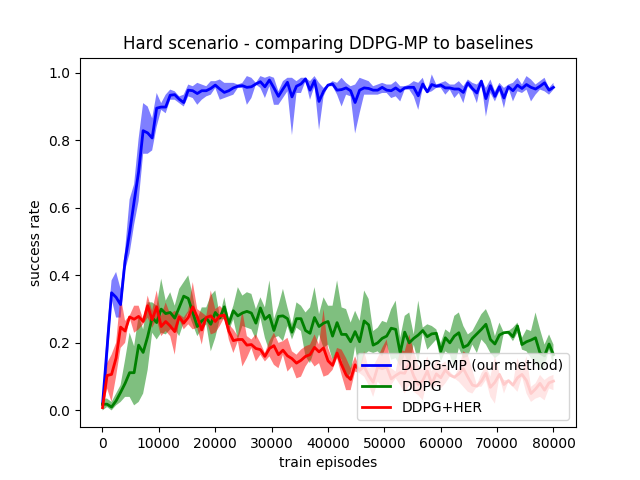}
\caption{}
\label{fig:hard-comparing-to-baseline}
\end{subfigure}~
\begin{subfigure}[t]{0.35\textwidth}
\includegraphics[width=\textwidth]{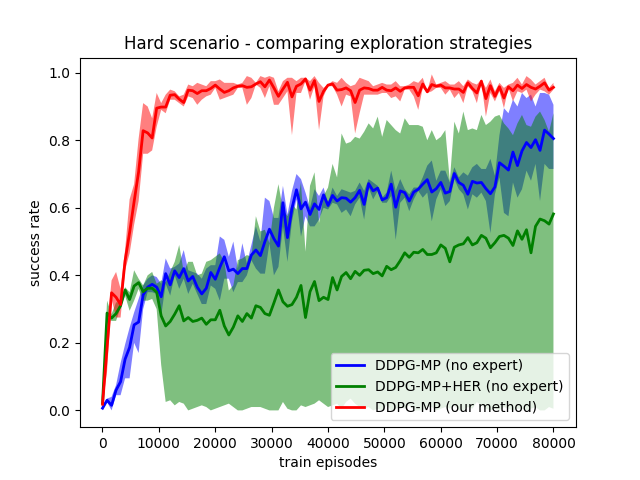}
\caption{}
\label{fig:hard-comparing-exploration}
\end{subfigure}
% \caption{Comparisons on \textit{hard} scenario. Left: baseline comparison. We compare DDPG and DDPG+HER in green and red, to DDPG-MP (our method) in blue. Right: comparing exploration strategies. We compare DDPG-MP with a vanilla DDPG exploration strategy (no expert) in blue, DDPG-MP with HER exploration strategy (DDPG-MP+HER) in green, and our exploration strategy utilizing expert demonstrations DDPG-MP (our method) in red.}
% \qquad
% \subfloat[][\textit{Vision} scenario - 1000 obstacle configurations selected randomly.]{
% \includegraphics[width=0.25\textwidth]{figures/scenario_vision.png}
% \label{fig:vision-scenario}
% }
% \qquad
% \subfloat[Subfigure 4 list of figures text][Subfigure 4 caption]{
% \includegraphics[width=0.4\textwidth]{figure4.jpg}
% \label{fig:subfig4}
% }
\caption{\textit{Hard} scenario results. \ref{fig:hard-comparing-to-baseline}: Comparison with baselines. We compare DDPG and DDPG+HER in green and red, to DDPG-MP (our method) in blue. 
% compares success rate compared to baseline models, and 
\ref{fig:hard-comparing-exploration}: Comparing exploration strategies. We compare DDPG-MP with a vanilla DDPG exploration strategy (no expert) in blue, DDPG-MP with HER exploration strategy (DDPG-MP+HER) in green, and our exploration strategy utilizing expert demonstrations DDPG-MP (our method) in red.
% compares exploration strategies. 
For both figures: each curve shows the average test success rate of 3 runs, with the upper and lower bounds the maximal and minimal values respectively. Each data-point is the count of successful work-spaces out of $200$ test work-spaces shown to the model during testing. }
\label{fig:success-plot-hard}
\end{figure*}

Next we evaluate DDPG-MP for NMP. We compare to two baselines: the original DDPG~\cite{lillicrap2015continuous}, and DDPG with HER~\cite{andrychowicz2017hindsight}, the current state-of-the-art in RL for learning goal-conditioned policies. 
% We show that  currently state-of-the-art approach of using DDPG+HER obtaining large success rate gains as demonstrated by 
Figures \ref{fig:success-plot-simple} and \ref{fig:success-plot-hard} plot the test success rate during training for the \textit{simple} and \textit{hard} scenarios. Tables \ref{tbl:hard-scenatio-success-rates} and \ref{tbl:vision-scenatio-success-rates} show the validation success rates for the \textit{hard} and \textit{vision} scenarios. 
Observe that DDPG-MP is the model with highest success rate in every experiment conducted, and on the \textit{hard} scenario, alternative approaches only reach $0.31$ success rate, while DDPG-MP reached a near perfect success rate of $0.97$. These results demonstrate the effectiveness of DDPG-MP in learning neural network policies for motion planning, especially when the problem is challenging in the sense that it requires navigating through narrow passages.

In the following we perform an ablation study to identify the components that make DDPG-MP so successful. 
% We also investigate the causes for these gains in an ablation study, designing experiments that test each of our contributions individually.

\subsection{Investigating the Model Based Actor Update}\label{sec:actor-stabilization}
In this experiment, we demonstrate that only changing the actor objective from Eq. \eqref{eq:ddpg-actor-update} to Eq. \eqref{eq:ddpg-mp-actor-update}, greatly improves the precision and speed-of-convergence for the agent. To establish this, we compare DDPG with a variant of DDPG-MP that does not use the expert demonstration data, but uses instead vanilla DDPG exploration. As shown in figure \ref{fig:success-plot-simple} and table \ref{tbl:simple-scenatio-success-rates},
DDPG trains on the full 64K work-spaces and reaches a validation success rate of $0.825$. 
Meanwhile, DDPG-MP needs in average of $33082.66$ episodes to reach a test success score of $1.0$ and stop, achieving a validation of $0.9936$.
% To clarify, the variant of DDPG-MP used here is not the full model as it lacks the exploration strategy. We used this limitation in order to isolate the contribution of the actor update rule change first, before understanding the benefit of our proposed exploration strategy which we show in the next experiment. 

% \begin{figure*}\label{fig:success-plot-hard}
% \centering
% \includegraphics[width=0.8\textwidth]{figures/success_plot_hard.png}
% \end{figure*}

\begin{table}[]
\begin{tabular}{lll}
\textbf{Model} & \textbf{Test} & \textbf{Validation} \\
\hline
DDPG                       & 865            & 0.82533 \\
DDPG-MP (no expert)        & \textbf{1.0}   & \textbf{0.99366} 
\end{tabular}
\caption{Success rates \textit{simple} scenario}
\label{tbl:simple-scenatio-success-rates}
\end{table}

\subsection{Investigating Exploration via Expert Demonstrations}\label{sec:harnessing-demonstrations}
The \textit{hard} scenario is a challenging motion planning environment to learn: the narrow corridor requires a very precise sequence of actions without much room for errors.
Moreover, since most work-space queries (which are presented randomly to the agent) do not include this portion of the state space, it is rare for the agent to see states in the corridor and is therefore prone to the problem of catastrophic forgetting.
Under regular RL exploration strategies this task is either impossible or requires a large amount of episodes which the exploration happens to give the correct sequence over and over.

The targeted exploration method via expert demonstrations of Section \ref{sec:targeted-exploration} 
% Our second algorithmic contribution which uses demonstrations from motion planners as an informed exploration strategy, 
is shown to be an effective solution to this problem.
In this section, we both measure the contribution of this strategy, and compare it to HER, a popular state-of-the-art exploration strategy. 
% We show, the not only our informed exploration strategy provides a significant boost in precision and sample efficiency, but it is far superior to HER, which despite the common belief is not well suited to this scenario.

We start by comparing the baselines DDPG with and without HER to DDPG-MP (full) as shown in Figure \ref{fig:hard-comparing-to-baseline}. We can see that DDPG-MP reaches a near-perfect validation success-rate of $0.9733$, while the DDPG and DDPG+HER only reach $0.318$ and $0.286$ each, demonstrating that the complete DDPG-MP algorithm is better in this scenario than both baselines.

Next, we would like to understand the contribution of using demonstrations to explore. For this purpose, in Figure \ref{fig:hard-comparing-exploration} we compare DDPG-MP with a vanilla exploration strategy (i.e., replacing the expert demonstrations exploration with the standard DDPG exploration) to DDPG-MP with a HER exploration strategy (i.e., replacing the expert demonstrations with HER) and the complete DDPG-MP algorithm. 
% each with its own "smart" exploration strategy (shown in figure \ref{fig:hard-comparing-exploration}).

Figure \ref{fig:success-plot-hard} shows the success rate over the number of training episodes, and Table \ref{tbl:hard-scenatio-success-rates} shows the average validation success-rate, both clearly show that DDPG-MP (full) is the best model.
Comparing DDPG-MP+HER (no expert) to DDPG-MP (no expert) we see that there is no benefit for using HER, and it even degrades the performance. We conclude that using HER is either not beneficial in this case, or requires more fine-tuning in order to make it work for these types of challenging motion planning scenarios. In Section \ref{sec:her-analysis} of the supplementary material we provide an in-depth analysis of why HER is not well suited for such motion planning scenarios.

Finally, we analyze our proposed exploration strategy by comparing DDPG-MP (no expert) to DDPG-MP (full). DDPG-MP (full) reaches a near-perfect success rate of $0.9733$ compared to $0.81933$ of DDPG-MP (no expert), and also learns 4X faster. 
% Moreover, figure \ref{fig:hard-comparing-exploration} shows that the speed of learning is also very different. DDPG-MP (no expert) still slowly improves until the end of the run after observing 80K episodes. On the other hand, DDPG-MP (full) reaches its peak performance with less than 20K episodes, making it both more accurate but also 4 times more sample efficient, 
These results clearly show that DDPG-MP's exploration strategy is beneficial.

% Figure \ref{fig:full-hard-trajectory} shows a work-space that requires a sequence of specific actions that cannot be learned without the proposed exploration policy.

\begin{table}[]
\begin{tabular}{lll}
\textbf{Model} & \textbf{Test} & \textbf{Validation} \\
\hline
DDPG                       & 0.36833           & 0.31766 \\
DDPG+HER                   & 0.3633            & 0.286 \\
DDPG-MP (no expert)        & 0.83              & 0.81933 \\
DDPG-MP+HER (no expert)    & 0.72833           & 0.68866 \\
DDPG-MP (full)             & \textbf{0.99 }    & \textbf{0.9733}
\end{tabular}
\caption{Success rates \textit{hard} scenario}
\label{tbl:hard-scenatio-success-rates}
\end{table}

% First we consider HER as a strategy. We notice that although HER provides a systematic way to overcome the sparse-rewards problem in RL, it does not provide a policy on how to cross a "narrow" states-corridors as in our scenario, which is exactly the reason why using HER will not reach a near perfect success rate.\AT{Can you describe how you obtained this explanation?} \TJ{remove passage}

\subsection{Vision scenario: Generalization to Unseen Obstacle Configurations}\label{sec:vision-generalization}
The previous two experiments showed that DDPG-MP can generalize to unseen work spaces, where the obstacles are fixed but the start and goal configurations are varied. 
In this experiment we also investigate the effect of changing the obstacle configurations -- can we train NMP models that generalize to previously unseen obstacle configurations?
To answer this question we use the \textit{vision} scenario, where with each work space, a visual input of the obstacles configuration is given to the model.
We hypothesize that if unseen work spaces share some similarities with the work spaces seen during training, DDPG-MP will be able to generalize and solve the unseen scenarios with high accuracy.

Table \ref{tbl:vision-scenatio-success-rates} shows that all models reach high success rates, with DDPG-MP getting the highest validation success rate of $0.93$. 
We note that the individual scenarios that comprise the \textit{vision} scenarios were selected from an easier distribution of obstacle configurations compared to the \textit{hard} scenario, and this may be the reason that models such as DDPG and DDPG+HER are almost on-par with DDPG-MP.

\begin{table}[]
\begin{tabular}{lll}
\textbf{Model} & \textbf{Test} & \textbf{Validation} \\
\hline
DDPG                       & 0.9233           & 0.89733 \\
DDPG+HER                   & 0.955            & 0.925 \\
DDPG-MP (full)             & \textbf{0.96166 }    & \textbf{0.93566}
\end{tabular}
\caption{Success rates \textit{vision} scenario}
\label{tbl:vision-scenatio-success-rates}
\end{table}

\subsection{Vision Scenario: Planning Times}\label{sec:planning-times}
Finally, we show that an NMP model trained with DDPG-MP has preferable running times to sample-based motion planners, for workspaces with an obstacle configuration not seen during training. To do this we compare the planning times it takes to produce trajectories for $100$ random workspaces with unseen obstacle configurations (from the \textit{vision} scenario test set). DDPG-MP takes $8.55$ seconds to compute, while it takes $50.93$ seconds for OpenRAVE's RRT motion planner -- a 6X speedup in favor of DDPG-MP. The experiment was conducted on a Ubuntu 16 desktop machine, with a 12-core Intel i7-8700k 3.7GHz CPU, 32GB RAM and NVIDIA GeForceGTX 1080Ti. We note that OpenRAVE uses C++ code, while our code runs in non-optimized python. We expect that with dedicated optimization these results would improve.

These planning times should be contrasted with the accuracy rate of the NMP approach. For a $93$\% success rate with a 6X speedup, a naive strategy that first tries to sample a plan using NMP, and if it fails (by running it through a collision checker) falls back to a conventional motion planner, a significant speedup on average would still be guaranteed. That said, we believe that smarter ways of combining the NMP with a conventional planner could be devised (e.g., along the lines of \cite{groshev2018learning}, which used an NMP as a search heuristic for discrete planning domains).

\section{Conclusion}
In this work we presented DDPG-MP: an RL based algorithm for training a neural motion planner. We showed that our method significantly improves the training time and final accuracy of the learned planner, compared to both imitation learning and state-of-the-art RL approaches.

We further showed that our approach can be used to train neural networks that predict accurate motion plans given an image of the obstacles in the domain, and do so considerably faster than conventional motion planners.

In this work, we focused only on the learning algorithm for training a neural motion planner. Our results pave the way for further investigations of neural network architectures for motion planning, and combinations of learning based approaches with classical methods for motion planning.
% a learning-based motion planning algorithm that uses RL in order to learn. We started with a motivation for using RL in the first place, by showing that IL methods do not consider important aspects of these problems, i.e states on the boundaries of the obstacles.

% Then, we analyzed our model and showed its advantages over previous work: it is more accurate and trains faster than other RL learning algorithms, and it plans faster than sample based motion planners on work-spaces with unseen obstacle configurations.

% \TJ{there are "future directions" below which I commented out. however i prefer to end on a positive note regarding our model and not emphasize its drawbacks...}
% It will be interesting to extend this work and investigate how using more immediate rewards (as suggested in section \ref{sec:ddpgmp-model-based-actor-update}) affects the stability of the algorithm.
% In this work we considered the dynamics of the world known and deterministic, since assumed that the time difference between actions is small enough and the low-level motor controllers are accurate. It would be interesting to see how this assumptions hold on a real robot.
% Another aspect of our model that needs to be considered for future implementation is that our model currently operates using a "global" context, i.e the image. In real-life sensors rarely provide such a global and accurate information and it would be intriguing to see if our method could instead handle a more "local" context and keep the success rate as high.

\pagebreak

\bibliographystyle{abbrvnat}
% \bibliography{references}
\bibliography{main}

\pagebreak

\appendix

\subsection{DDPG-MP Algorithm}\label{sec:algorithm}
\begin{algorithm}[htp]
  \SetAlgoLined\DontPrintSemicolon
  \SetKwFunction{algo}{DDPG-MotionPlanner}
  \SetKwFunction{collectEpisodes}{CollectEpisodes}
  \SetKwFunction{append}{Append}
  \SetKwFunction{getDemonstration}{GetDemonstration}
  \SetKwFunction{scoreReward}{ScoreReward}
  \SetKwFunction{addToReplayBuffer}{AddToReplayBuffer}
  \SetKwFunction{modelUpdate}{ModelUpdate}
  \SetKwProg{myalg}{Algorithm}{}{}
  \myalg{\algo{} }{
  \nl Load pre-trained reward $\widetilde{r}$ and transition-classification $\widetilde{p}$ models\;
  \nl Randomly initialize critic network $\hat{Q}\left(w,c,a|\theta^{\hat{Q}}\right)$ and actor $\pi\left(w,c|\theta^{\pi}\right)$ with weights $\theta^{\hat{Q}}$ and $\theta^{\pi}$ \;
  \nl Initialize target network $\hat{Q}'$ and $\pi'$ with weights $\theta^{\hat{Q}'}\leftarrow \theta^{\hat{Q}}$ and $\theta^{\pi'}\leftarrow \theta^{\pi}$\;
  \nl Initialize replay buffer $R$ \;
  \For{$iteration: 1...M$}{
    \nl $episodes\leftarrow $ \collectEpisodes{} \;
    \For{$e$ in $episodes$}{
        \If{$e.status$ is $failed$}{
            \nl $d\leftarrow $\getDemonstration{$e$} \;
            \nl $episodes.$ \append{ \scoreReward{$d$} }
        }
    }
    \nl \addToReplayBuffer{$R$,$episodes$} \;
    \For{$update: 1...U$}{
        \nl \modelUpdate{$R$, $\widetilde{r}$, $\widetilde{p}$, $\hat{Q}$, $\pi$, $\hat{Q}'$, $\pi'$}
    }
  }
  }
  { }
  \setcounter{AlgoLine}{0}
  \SetKwProg{myproc}{Procedure}{}{}
  \myproc{\modelUpdate{$R$, $\widetilde{r}$, $\widetilde{p}$, $\hat{Q}$, $\pi$, $\hat{Q}'$, $\pi'$}}{
  \nl Sample a random minibatch of $N$ transitions $(w,c_i,a_i,r_i,c_{i+1})$ from $R$ \;
  \nl $y_i\leftarrow r(w, c_i, c_{i+1}, a_i)$ \;
  \If{$(w,c_i,c_{i+1})\in T_{free}$}{
    $y_i\leftarrow y_i + \gamma \hat{Q}'(w,c_{i+1}, \pi'(w, c_{i+1}))$
  }
  \nl Update critic by minimizing the loss $\frac{1}{N}\sum_i{\left(y_i - \hat{Q}(w,c_i,a_i|\theta^{\hat{Q}})\right)^2}$ \;
  \nl Update the actor policy using the sampled policy gradient:
  \begin{equation*}
      \frac{1}{N}\sum_i{\bigtriangledown_\theta \left(\widetilde{r}(w,c_i,a_i) + \gamma \widetilde{p}(w,c_i,a_i)\hat{Q}(w,c_i,a_i)\right)}
  \end{equation*} \;
  \nl Update the target networks:
  \begin{align*}
  &\theta^{\hat{Q}'}\leftarrow \tau {\theta^{\hat{Q}}} + (1-\tau) \theta^{\hat{Q}'} \\
  &\theta^{\pi'}\leftarrow \tau \theta^{\pi} + (1-\tau) \theta^{\pi'}
  \end{align*} \;
  }
  \caption{DDPG for Motion Planning}
  \label{alg:ddpg-mp}
\end{algorithm} 

In this section we present the full DDPG-MP algorithm shown in Algorithm \ref{alg:ddpg-mp}.
Our extensions to DDPG can be found at:
\begin{enumerate}
    \item Our actor update from Section \ref{sec:ddpgmp-model-based-actor-update} Equation (\ref{eq:ddpg-mp-actor-update}) is found in line 4 of the \modelUpdate method.
    
    \item The exploration via expert demonstration of Section \ref{sec:targeted-exploration} is shown on lines 6 and 7 of  \algo .
\end{enumerate}

\subsection{Experiments Settings and Parameters}\label{sec:experiments-and-parameters}
In this section we describe the technical details we used for our scenarios in Section \ref{sec:results}, as well as the hyper-parameters used for training the models. 

We start by specifying the environment parameters used by our OpenRAVE~\cite{diankov2008openrave} simulator. These values remain the same for all experiments for both the IL and RL agents. The simulator acts in joint movements the size of $0.025$ in the $L2$ norm (re-scaled the input action if required), while collision is tested on a grid spaced by $0.001$ intervals. 
Since it is unlikely that any agent would be able to reach the goal pose exactly, we allow that the finishing goal pose of the trajectory to have a distance smaller than $0.04$ between itself and the goal pose $g$.

We now discuss how to select work spaces that are suitable for training. A work space $W=(F,c_0,g)$ must be challenging in order for NMPs trained on a corpus of such work spaces to have the ability to solve complex work spaces. In our work we decided that a work space can be inserted into our corpus of training data if exists an obstacle with center $o$ such that $o$ is between the start and goal poses, i.e $\|o-FK(c_0)\|_2 \le \|g-FK(c_0)\|_2$ and $\|o-g\|_2 \le \|g-FK(c_0)\|_2$. This formulation only selects demonstrations that show movement from one side of an obstacle to another. We note that in the \textit{hard} and \textit{vision} domains, many work spaces require that the agent move across more than one obstacle.

When learning, each agent trains in an iterative process according to its structure and target functions, IL by epochs, and RL using noisy episode roll-outs. Between iterations, the model is constantly tested on work-spaces from the \textbf{test} set and the current test success-rate is evaluated. If the currently evaluated score is the best seen so far the model is saved as \textit{best iteration}.
After training, the best iteration model is re-evaluated using $1000$ new unseen work-spaces, and this is reported as the validation success-rate. 
We do this to make sure that the reported scores are not high due to randomness but actually reflect the performance of the agent.
We note that the information gathered during testing and validation do not affect the training, namely these observed transitions are not added to any data-set the agents use for training.

The RL setting requires the following parameters: the future-rewards coefficient $\gamma$ is set to $0.99$. The reward at each time-step is combined from both $r^T_t$ as described in Eq. \ref{eq:reward}, and a domain reward. 
Our domain reward, penalizes the agent when moving a joint beyond the joint's limit. The distance wasted is multiplied with a coefficient of $0.05$ and this is the additional penalty $r^D$.

All versions of DDPG and DDPG-MP use the following parameters: the agent collects $16$ episodes, after which $40$ model-update operations are executed sequentially (using different batches from the replay buffer). 
The learning continues until 64K, 80K and 160K episodes are seen for the \textit{simple}, \textit{hard} and \textit{vision} scenarios respectively.
The batch size is $512$, while the replay buffer size is $1000000$, the parameter $\tau$ which updates the target networks towards the online networks is set to $0.95$.

The exploration strategies' parameters are as follows: for \textit{DDPG} agents (with and without HER) and for \textit{DDPG (no expert)} we use the same default random noise as in the original DDPG paper~\cite{lillicrap2015continuous}, completely random exploration occurs with $0.2$ probability, and $0$-mean Gaussian noise with $0.05$ std is added to every (training) action. 
Our model DDPG-MP which is not require extensive random exploration instead uses $0.02$ chance for a random action, and the Gaussian noise std is $0.005$. To refute the claim that the change of parameters is the reason our model learns better, we also ran tests where \textit{DDPG} used this set of exploration parameters, and sure enough, DDPG discovered almost nothing and the success-rate remained close to zero for the entire training process.

Regarding the HER~\cite{andrychowicz2017hindsight} parameters, we tried to take the default values, using the "future" strategy, and $k=4$ as recommended. This worked on the \textit{hard} scenario, however in the \textit{vision} scenario, we needed to reduce $k$ to $1$, since the success rates of model with $k>1$ was dropping to zero after only a few training cycles without recovering.

DDPG-MP's exploration policy utilizes motion plans for failed work-spaces. We found that instead of taking all the expert demonstrations that correspond to failed episodes, if we limit to a maximum of $8$ episodes per update (out of $16$) we get better performance.

In the next part we describe implementation details regarding the neural network models.
All our networks start by concatenating the current joints $c_t$ to the information within the context vector $w$, and using it as input to subsequent layers.
Meaning, in the \textit{simple} and \textit{hard} scenarios where there is no visual information, the result is just the goal pose $g$ concatenated with $c_t$.
The \textit{vision} scenario also contains visual information that first needs to be processed. 
We use a visual component (described below) that maps the input image $I_F$ describing the obstacles configuration, to a vector. This vector, gets concatenated to $c_t$ and the goal pose $g$.

The visual component is based on the DQN~\cite{mnih2015human} design with less layers: we use two convulotional layers with $32$ and $64$ filters respectively, with kernel sizes of $8$ and $4$, and strides of $4$ and $2$. The result is flattened and goes through 2 fully-connected layers with $512$ hidden layers.
All layers use the ReLU activation function~\cite{DBLP:journals/corr/abs-1803-08375}. The input for the visual component, $I_F$, is a $55\times 111$ grey-scale image.

The policy network (or actor) is a 4 hidden-layer NN with $200$ neurons in each layer and Elu activation units~\cite{DBLP:journals/corr/ClevertUH15}, with the final layer being a tanh activation. This is the same for all IL and RL agents.
To make the network output a direction vector, the result of the tanh layer is normalized to always have a norm of 1.
We found that in order to keep the learning stable, $L2$ regularization with coefficient of $1.0$ should be used on the input to the tanh function (the pre-activation values).

The critic has 3 layers with $400$ neurons each and the output of the third layer is concatenated the action input. Then 4 more layers with $400$ neurons each follow.
The critic also uses Elu activations, except for the output that is linear, and we found that regularizing the critic weights with coefficient $10e-7$ works best.

As for the RL optimization, we used the Adam optimizer~\cite{DBLP:journals/corr/KingmaB14}, and both actor and critic networks use a learn rate of $0.001$. We also found that using a gradient limit of $1.0$ was important to keep the gradients from exploding.

The imitation learning agents (both behavioral cloning and DAgger) use the same actor network structure as the RL agents. We train each agents for 100 epochs (regardless of the data-set size). These agents also use the Adam optimizer, but with a learning rate of $0.0001$ and a gradient limit of $5.0$.

All our agents were implemented using Tensorflow~\cite{tensorflow2015-whitepaper}, which is a python deep-learning with auto-differentiation package.

\subsection{Training IL Agents in the \textit{Simple} scenario}\label{sec:trainig-IL-agents}
In this section we elaborate on the preliminary tests regarding the usage of Imitation Learning in our settings. 
The goal of this experiment was to evaluate the success rate that IL agents can achieve and understand their data-set requirements, since in the motion planning settings obtaining motion-plans is costly. 
We found that while consuming a large amount of data, these agents do not reach the success rate expected from motion-planners. 
For more information regarding the reasons why IL is not well-suited see Section \ref{sec:IL-vs-RL} and Figure \ref{fig:data_distribution} in the main text.

Since the motion planning simulator described in Section \ref{sec:experiments-and-parameters} requires a direction vector, we use the cosine-distance as a loss function (the cosine distance is measured between the predicted motion direction to the direction of the actual motion seen in training).

This experiment evaluates the success rate of both, Behavioral-Cloning and DAgger. 
For Behavioral-Cloning, even though we had 160K episodes of demonstrations, we also wanted to evaluate how the success improves as a function of the number of examples seen during training. 
Table \ref{tbl:il-success-rate-by-data-set-size} shows that dependency, and we note that the it appears that Behavioral-Cloning agents reach a maximal validation success-rate of only about $0.79$, and that additional data does not improve the results.

DAgger~\cite{ross2011reduction}, is an iterative method that uses an online expert to score states visited during training in order to avoid the distribution miss-match problem.
We note, that in the motion planning case this approach is costly, since it requires the motion planner to plan an entire motion plan for every state seen during training (a single train episode contains hundreds of state visitations). 
This makes testing DAgger even more challenging since demonstrations are not computed in advance like in the IL case, but rather during the training process. 
We therefore relaxed this constraint and require the motion planner to plan for only the last 5 states in a failed work-space query.

Because of the expensive data requirements, we seed DAgger with 14K motion-plan demonstrations, and after every one of its 100 training epochs, DAgger gets 500 episodes that it can score using the policy mentioned above. Training for 100 epochs allows DAgger to see 64K episodes.
With this configuration DAgger reaches a validation success rate of $0.8$, and we found that either adding more seeded motion-plans or online demonstrations does not affect the success rate. 
Meaning that similarly to Behavioral-Cloning the performance of DAgger is also capped albeit a bit higher.

\begin{table}[]
\begin{tabular}{ll}
\textbf{Number of episodes seen} & \textbf{Validation success rate} \\
\hline
10K                       & 0.725\\
20K                       & 0.67\\
40K                       & 0.73\\
80K                       & 0.76\\
120K                       & 0.789\\
160K                       & 0.775\\
\end{tabular}
\caption{Validation success rates Behavioral-Cloning by number of motion plans demonstrations seen}
\label{tbl:il-success-rate-by-data-set-size}
\end{table}

\subsection{Reward Models for Motion Planning: Design and Training}\label{sec:reward-models}
In this section we describe the structure of our reward model $\widetilde{r}$, the transition-termination model $\widetilde{p}$, as well how to train and test them.
To utilize the full information available in a demonstration of a motion planning step $(w,c_t, c_{t+1})$ we can exploit two labels: the transition class ($T_{free}$, $T_{goal}$ or $T_{col}$) and the reward value itself $r_t(w,c_t, c_{t+1})$.
This allows us to jointly train the models $\widetilde{r}$ and $\widetilde{p}$ in a single network.

We start by building and training a model for $\widetilde{p}$: this is a network with input $(w,c_t, c_{t+1})$ (inline with our other models as described in section \ref{sec:experiments-and-parameters}). 
The network outputs 3 hidden neurons that are the logits of the transition-class label: $l_{free}$, $l_{goal}$ and $l_{col}$ matching $T_{free}$, $T_{goal}$ or $T_{col}$. 
We then use the softmax function to get probabilities for each transition class, denoted by $\widetilde{p}_{free}$, $\widetilde{p}_{goal}$ and $\widetilde{p}_{col}$.

We train this network with a cross-entropy loss with the label being the true transition class label. 
The architecture of this network is simple: the input, is followed by 4 fully-connected hidden-layers with 100 neurons and Elu~\cite{DBLP:journals/corr/ClevertUH15} activations each.

To compute $\widetilde{p}$, we can either take the $1-\widetilde{p}_{free}$ (probability of not ending in termination according to the softmax). Or we can first decide on the most probable between $l_{goal}$ and $l_{col}$ i.e $l_{termination} = \max (l_{goal}, l_{col})$, and consider its softmax value between $l_{termination}$ with $l_{free}$. Empirically we found no difference between the two.

After this training process, we have a network that assigns probabilities to class transitions, we can train the predictor $\widetilde{r}$ for the reward. The input for this network is both (1) the same input as $\widetilde{p}$'s and (2) a 3 dimensional vector that is the probability for each transition class.
The output for this network is a single scalar with a loss function of Mean-Squared-Error (MSE).We can use any network architecture to process this input, but in our experiments we simply used a linear classification of the inputs.

We notice that the above reward model allows us to use the output of the transition-labeling network i.e. $\widetilde{p}_{free}$, $\widetilde{p}_{goal}$ and $\widetilde{p}_{col}$ as an input to $\widetilde{r}$. This is what allows us to use this model in update rule in Equation (\ref{eq:ddpg-mp-actor-update}) where the transition-label itself is not known in advance.

We train these model using the Adam optimizer~\cite{DBLP:journals/corr/KingmaB14} with a learning rate of $0.001$ a gradient limit of $5.0$ and a $L_2$ weight decay with coefficient $0.0001$. The batch size for the \textit{simple} and \textit{hard} scenarios is 10k, and for the \textit{vision} scenario is 2K.

\textbf{An important note about transition label imbalance:} Empirically, in most motion planing scenarios, the number of transitions that are free $(w,c_t,c_{t+1}) \in T_{free}$ greatly outnumbers the other two classes. In all of our examples, transitions in $T_{free}$ make up of more than $0.98$ of the total transitions. This makes the other two classes very sparse, which make it hard to find a good solution (the trivial predictor that always predicts $T_{free}$ has an accuracy of more than $0.98$).
To handle this issue we over-sampled the classes $T_{goal}$ and $T_{col}$ in each batch, making all classes equally proportional. 

\subsection{Analysis on HER Performance Degradation}\label{sec:her-analysis}
In this section we attempt to give empirical evidence to explain why HER, is not a good exploration strategy to use on scenarios with narrow state passages. Our main observation is that HER has a tendency to amplify jittery movements.

An implicit assumption in HER is that every movement in an episode is a correct step towards the final state of that episode, however this assumption does not hold for work-spaces that contain unexplored narrow passages.
Without a good exploration policy to guide the agent through the passage, the agent often becomes "hesitant" to enter, and starts going back-and-forth near the entrance to the narrow region as it is trapped in a local minima, thus making no progress towards the goal on the other side of the passage.
This phenomenon would only increase by HER, since the shaking actions are not only non-ideal progress even towards the final state, they are often even in the opposite direction.

Consider the results from Section \ref{sec:harnessing-demonstrations} of DDPG compared to DDPG+HER, the latter shows a consistent faster increase in test success-rate during the first 10K episodes of training, which aligns with the goal of HER to make rewards less sparse.
However, as the agents start to consider the passages more and more as "danger zones", the jittering phenomena described above becomes more and more amplified, and DDPG surpasses DDPG+HER on average in the 25K episodes mark.
We refer the reader to the uploaded movie which also shows the jittering phenomena with HER.

\end{document}